%% file: main.tex
\documentclass[letterpaper, 10 pt, conference]{ieeeconf}  

\IEEEoverridecommandlockouts                              
\usepackage{algorithm}
\usepackage{algpseudocode}
\usepackage{balance}
\usepackage{epstopdf}

\usepackage{graphicx}
\graphicspath{ {./images/} }
\usepackage{subfig}

\usepackage{subcaption}
\usepackage{amsmath,amssymb,amsfonts}
\usepackage{xcolor}
\usepackage{bm}

\usepackage{amsmath}

\usepackage[colorinlistoftodos]{todonotes}

\title{\LARGE \bf
Dynamic Walking on Highly Underactuated Point Foot Humanoids: Closing the Loop between HZD and HLIP}

\author{Adrian~B.~Ghansah$^{1}$, Jeeseop~Kim$^{2}$, Kejun~Li$^{3}$, Aaron~D.~Ames$^{1,2}$
\thanks{$^{1}$A. B. Ghansah, and A. D. Ames are with the Department of Control and Dynamical Systems, California Institute of Technology, Pasadena, CA 91125, USA, {\tt\small \{aghansah, ames\}@caltech.edu}\newline
$^{2}$J. Kim, and A. D. Ames are with the Department of Mechanical and Civil Engineering, California Institute of Technology, Pasadena, CA 91125, USA, {\tt\small \{jeeseop, ames\}@caltech.edu}\newline
$^{3}$K. Li is with the Department of Computation and Neural Systems, California Institute of Technology, Pasadena, CA 91125, USA, {\tt\small \{kli5\}@caltech.edu}
}%
}

\begin{document}

\maketitle
\thispagestyle{empty}
\pagestyle{empty}

\input{00_abstract}

\input{01_introduction}

\input{02_background}

\input{03_method}

\input{04_application}

\input{05_experiments}

\input{06_conclusion}


\bibliographystyle{IEEEtran}
\bibliography{99_references}

\end{document}

%% file: 00_abstract.tex
\begin{abstract}
    Realizing bipedal locomotion on humanoid robots with point feet is especially challenging due to their highly underactuated nature, high degrees of freedom, and hybrid dynamics resulting from impacts.  
    With the goal of addressing this challenging problem, this paper develops a control framework for realizing dynamic locomotion and implements it on a novel point foot humanoid: ADAM.  To this end, we close the loop between Hybrid Zero Dynamics (HZD) and Hybrid linear inverted pendulum (HLIP) based step length regulation.  To leverage the full-order hybrid dynamics of the robot, walking gaits are first generated offline by utilizing HZD. These trajectories are stabilized online through the use of a HLIP based regulator. Finally, the planned trajectories are mapped into the full-order system using a task space controller incorporating inverse kinematics.
    The proposed method is verified through numerical simulations and hardware experiments on the humanoid robot ADAM marking the first humanoid point foot walking.  Moreover, we experimentally demonstrate the robustness of the realized walking via the ability to track a desired reference speed, robustness to pushes, and locomotion on uneven terrain.   
\end{abstract}

%% file: 01_introduction.tex
\section{Introduction}\label{sec:introduction}

A central and long-standing goal of humanoid robots has been the realization of dynamic locomotion that will enable these robots to locomote in natural environments.  
This has recently become even more important as the number of humanoid robots have grown, being developed in both academic environments and by industry.  There have been great successes in this regard---humanoids that can walk in outdoor environments, natural and human-like multi-contact locomotion, and efficient walking.  Yet challenges remain to achieve the robustness needed for humanoid robots to navigate complex human-centered environments.  This paper addresses arguably the hardest form of bipedal locomotion: point contact walking.  This results in highly underactuated dynamics without even a passive foot to assist with stabilization.  While this is not the final form a humanoid robot would take---feet are without a doubt beneficial---studying this harder problem gives insight into control strategies that will result in more dynamically stable and robust locomotion on humanoids---even those with feet. 



One approach that has proven successful in realizing dynamic locomotion on highly underactuated bipedal robots is the Hybrid Zero Dynamics (HZD) method\cite{westervelt2003hybrid}. This framework captures the hybrid nature of walking by using hybrid dynamic models that consists of domains with the full order continuous dynamics connected via discrete transitions caused by the impact dynamics. Furthermore, by enforcing virtual constraints through feedback control, it is possible to generate provable stable walking \cite{grizzle2014models,westervelt2018feedback,ames2014rapidly}. Since the full-order dynamics are leveraged, this approach is platform agnostic and it has successfully been demonstrated on a number of robotic platforms including various bipeds \cite{reher2016realizing, reher2019dynamic}, robotic assistive devices \cite{tucker2021preference, li2022natural} and for different behaviors such as heel-to-toe walking and running \cite{ma2016efficient}.  Yet HZD is computationally expensive (gaits must be computed offline), and therefore it does not allow for the re-planning of gaits online in a step-to-step fashion. 
As a result, in practice, these gaits are often deployed with Raibert style foot placement regulators \cite{reher2021dynamic}.


\begin{figure}
    \centering
    \includegraphics[width=\linewidth]{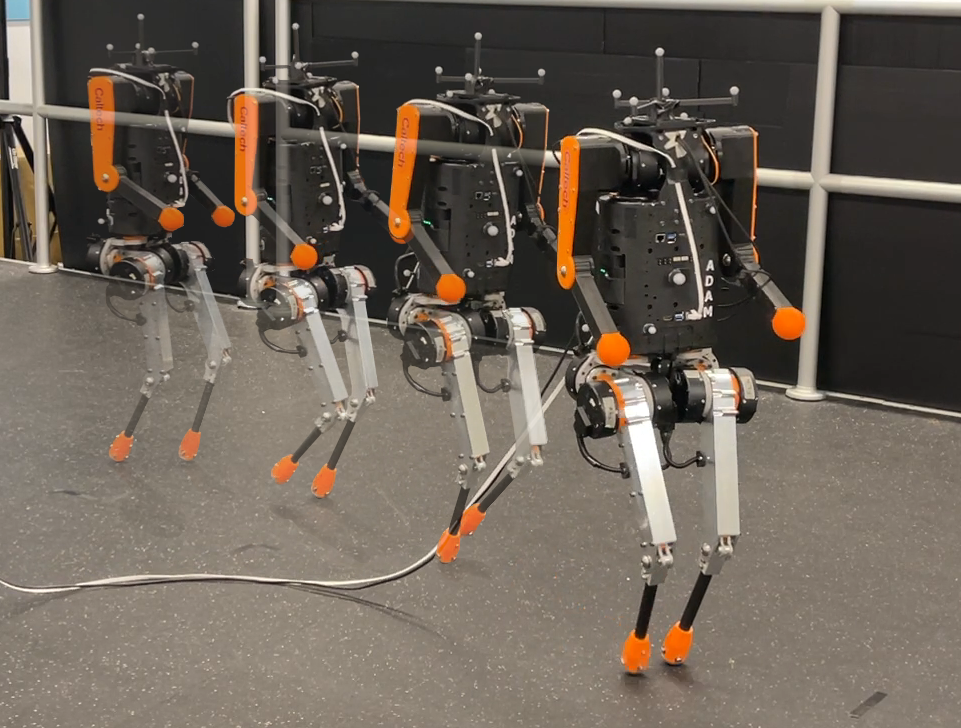}
    \caption{Snapshots of ADAM tracking the nominal HZD gaits with a stabilizing HLIP controller.}
    \label{fig:hero}
    \vspace{-1.5em}
\end{figure}
\begin{figure*}[h]
    \centering
  \includegraphics[width=0.97\textwidth]{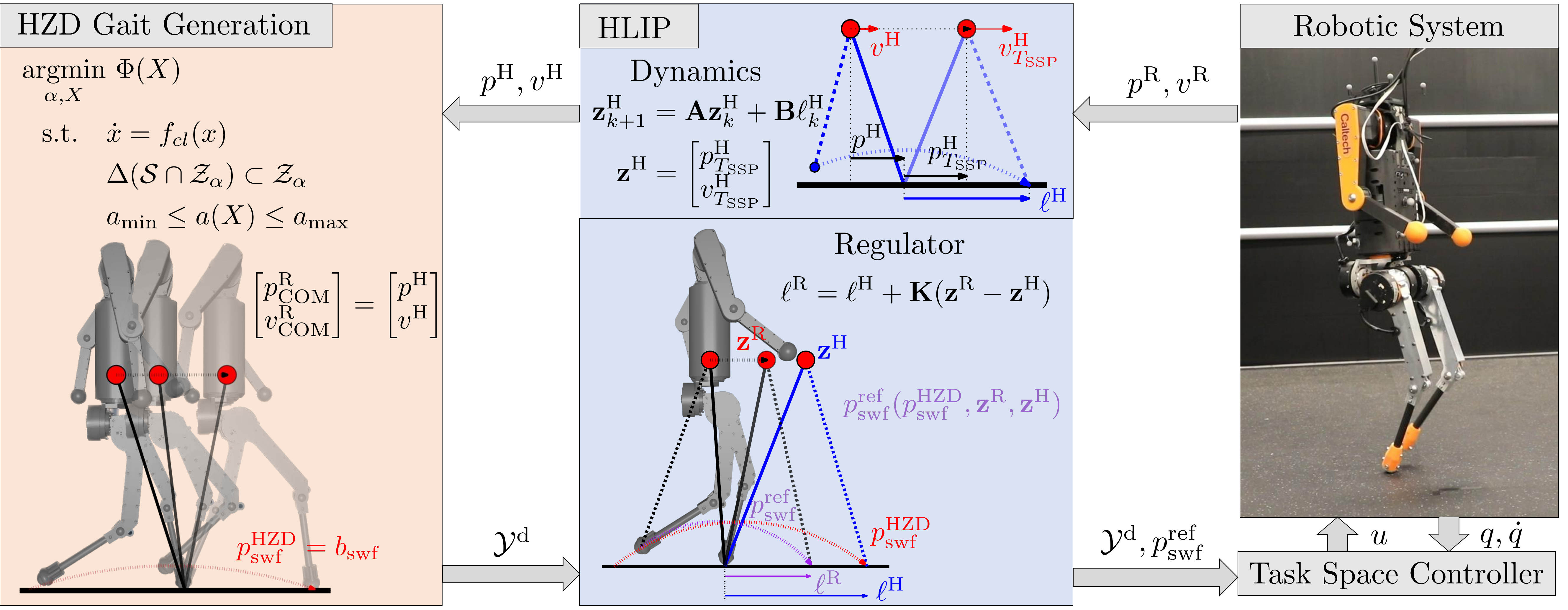}
  \caption{An overview over the control framework showing the closed loop interaction between the HZD gait generation and the HLIP regulator.}
  \label{fig:control_architecture}
  \vspace{-1.8em}
\end{figure*}

As a counterpart to the full-order model based HZD approach, reduced-order models (ROMs) can be easily integrated into an online planning and feedback control framework. 
Trajectory planning, leveraging zero-moment-point (ZMP) based on a linear inverted pendulum (LIP) model, has been widely applied to quasi-static humanoid walking \cite{kajita2002realtime, kajita2003biped, englsberger2011bipedal}. However, the LIP model has not been studied actively in the context of underactuated bipedal robots due to the gap between the LIP dynamics and the inherent nature of underactuated systems.
To improve upon this, several popular ROMs for bipedal robots, including the spring-loaded LIP (SLIP) model \cite{blickhan1989spring} and the hybrid LIP (HLIP) model \cite{xiong2019orbit}, have demonstrated effective trajectory planning and control for locomotion across various underactuated bipedal robot platforms \cite{da20162d, xiong20223}.
Despite these advancements, trajectory planning using ROMs presents its own set of challenges. Notably, there is no assurance that the full-order dynamics will align with the simplified dynamics when following the reduced-order trajectories, i.e., that the reduced-order model generates dynamically feasible trajectories.

In this work, we combine the benefits from the generality of the HZD framework and the computational benefits of HLIP to stabilize severely underactuated bipedal robots---namely, humanoids with point feet.  
This is achieved through a control framework that generates stable and robust locomotion on point foot humanoid robots by closing the loop between the full-order hybrid model gait generation (via HZD) and the reduced-order model gait stabilization (via HLIP). A high-level overview of this approach is shown in Fig. \ref{fig:control_architecture}.
More concretely, we are able to close the loop between HZD and HLIP by (1) accounting for the ROM (HLIP) dynamics when generating the gaits offline using HZD and (2) stabilizing these HZD gaits online using an HLIP-based footstep algorithm.  This framework is verified in both a high-fidelity simulation and experimentally on a novel point foot humanoid: ADAM.  
To experimentally demonstrate the robustness of the proposed approach, we consider the ability of the point foot humanoid ADAM to track a desired reference walking velocity, recover from unplanned pushes, and locomote over rough terrain.  
These experiments mark what the authors believe to be the first example of point foot walking on a humanoid robot.




The paper is organized the following way. In section \ref{sec:background} we present how trajectory optimization and the HZD framework can be used to generate gaits, and provide preliminaries on the HLIP model. Section \ref{sec:gait_stabilization} illustrates how the HLIP model is embedded into the HZD gait generation, and how the HLIP regulator modifies the nominal HZD outputs online. Section \ref{sec:application} shows application of the control framework to the humanoid robot ADAM \cite{ghansah2023humanoid}, and section \ref{sec:experimental} presents the simulation and hardware experimental results. Finally, in section \ref{sec:conclusion} we provide concluding remarks.



%% file: 02_background.tex
\section{Background}\label{sec:background}

The dynamics of a walking robot consist of continuous phases connected by discrete impacts. This combination of continuous and discrete dynamics makes the system a hybrid system. In section \ref{subsec:background:hzd_framework}, we outline the core principles behind the HZD framework, which has proved itself effective to control hybrid systems. In section \ref{subsec:background:trajectory_optimization}, we present how trajectory optimization can be used to generate nominal gaits based on the full-order hybrid model. Following this, we include preliminaries on the HLIP model \cite{xiong20223} in section \ref{subsec:background:hlip} which is the ROM we use to stabilize the nominal HZD gaits.

\subsection{Hybrid Zero Dynamics Framework}\label{subsec:background:hzd_framework}

Consider a robotic system with configuration coordinates $q \in \mathcal{Q} \subset \mathbb{R}^n$, and let the full system state be given by $x = (q, \dot{q}) \in \mathcal{X} \subset  \mathsf{T}\mathcal{Q}$. The system can be represented in standard form as a second-order mechanical system 
\begin{equation}
    D(q) \ddot{q} + H(q, \dot{q}) = Bu,
    \label{eq:EL}
\end{equation}
where $D(q) \in \mathbb{R}^{n \times n}$ is the inertia matrix, $H(q, \dot{q}) \in \mathbb{R}^n$ is the drift vector, $B \in \mathbb{R}^{n \times m}$ is the actuation matrix, and $u \in \mathcal{U} \in \mathbb{R}^m$ is the system input. By expressing the dynamical system in state-space form it can be written as
\begin{equation}
    \dot{x} = f(x) + g(x)u.
\end{equation}

Since we are considering a system without compliance, we limit our scope to a single-domain single-edge hybrid system. We assume that the transition from the current single support phase (SSP) to next happen instantaneously. Let $p_{\textnormal{swf},z} : \mathcal{Q} \rightarrow \mathbb{R}$ denote the swing foot height above the ground. The set of admissible states within the domain are then given by
\begin{equation}
    \mathcal{D} := \{ (q, \dot{q}) \in \mathcal{X} \ | \ p_{\textnormal{swf},z}(q) \geq 0 \} \ \subset \mathcal{X}.
\end{equation}

When the swing foot impacts with the ground, the system goes through a discrete jump and transitions to the next domain. The states corresponding to the edge are denoted as the switching surface $\mathcal{S} \subset \mathcal{D}$ which is defined as
\begin{equation}
    \mathcal{S} := \{(q, \dot{q}) \in \mathcal{X} \ | \ p_{\textnormal{swf},z}(q) = 0, \ \dot{p}_{\textnormal{swf},z}(q, \dot{q}) < 0\}.
\end{equation}
This change in states can be obtained by using the reset map $\Delta : \mathcal{S} \rightarrow \mathcal{X}$, defined as
\begin{equation}
    x^+ = \Delta(x^-), \quad x^- \in \mathcal{S}
    \label{eq:reset_map}
\end{equation}
where $x^-$ and $x^+$ refer to the pre- and post-impact states, respectively.

The continuous dynamics within a domain can be combined with the reset map from \eqref{eq:reset_map} to form a single-domain hybrid control system given by
\begin{equation}
    \mathcal{HC} = 
    \begin{cases}
        \dot{x} = f(x) + g(x)u \quad & x\notin \mathcal{S} \\
        x^+ = \Delta(x^-) & x^- \in \mathcal{S}
    \end{cases}
    \label{eq:hzd_single_domain}
\end{equation}
The key concept of the HZD framework is to drive the hybrid dynamical system from \eqref{eq:hzd_single_domain} to evolve on a lower dimensional manifold via the design of virtual constraints. Let the actual outputs be denoted by $\mathcal{Y}^a(q)$, and let the desired outputs be denoted by $\mathcal{Y}^d(\tau(q), \alpha)$. Through virtual constraints $\mathcal{Y}_{\alpha}: \mathcal{Q} \rightarrow \mathbb{R}^m$, we can shape the behavior of the outputs
\begin{equation}
    \mathcal{Y}_{\alpha}(q) := \mathcal{Y}^a(q) - \mathcal{Y}^d(\tau (q), \alpha).
    \label{eq:output_constraint}
\end{equation}
Here $\alpha \in \mathbb{R}^{m \times (b+1)}$ is a collection of Bézier coefficients, where  $b \in \mathbb{N}_{\geq 0}$ denotes the order of the Bézier polynomials.

When all the virtual constraints and their derivatives are driven to zero, the system evolves on the zero dynamics surface, which is defined as
\begin{equation}
    \mathcal{Z}_{\alpha} := \{ x \in \mathcal{D} \ | \ \mathcal{Y}_{\alpha}(q) = 0, \ \dot{\mathcal{Y}}_{\alpha}(q) = 0 \}.
    \label{eq:zero_dynamics_surface}
\end{equation}
By using a feedback controller $u^*(x)$ it is possible to drive the outputs to zero exponentially on the continuous domains, which results in a closed-loop system given by $\dot{x} = f_{cl}(x) := f(x) + g(x)u^*(x)$. Furthermore, to ensure that the hybrid system as a whole remains stable, we must make sure that the outputs remain zero also through the impacts, which can be captured through the following impact-invariance condition:
\begin{equation}
    \Delta (\mathcal{S} \cap \mathcal{Z}_\alpha) \in \mathcal{Z}_\alpha.
    \label{eq:HZDcond}
\end{equation}

\begin{figure}[t!]
    \centering
    \subfloat[\centering Nominal state]{{\includegraphics[width=3cm]{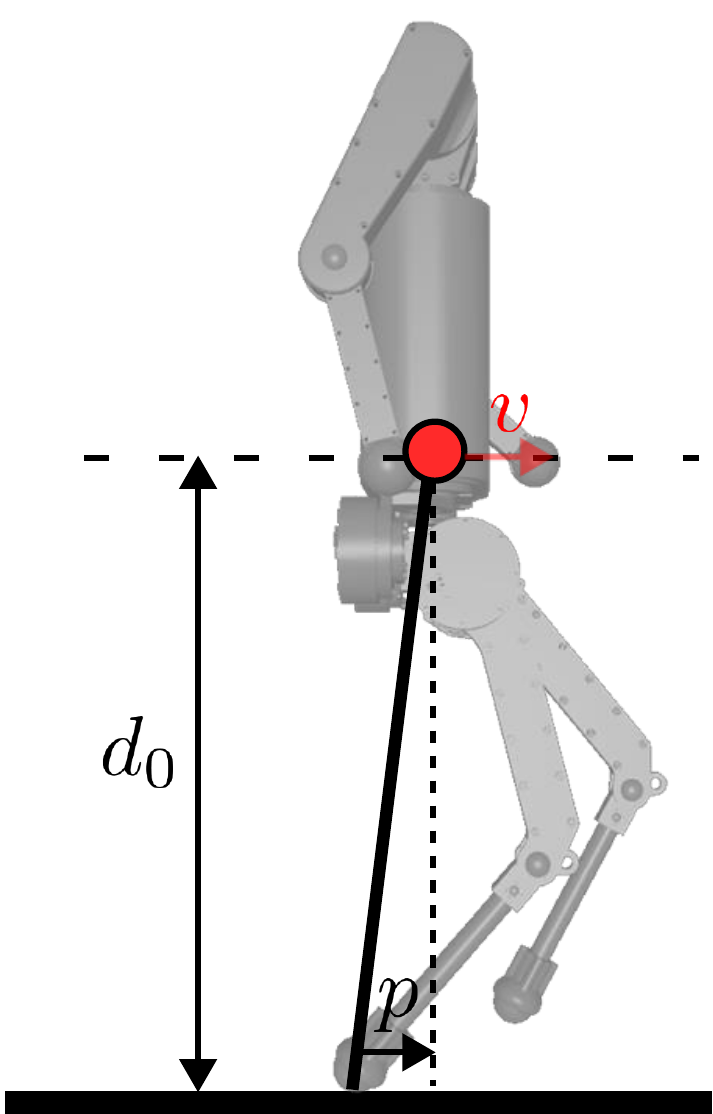} }}%
    \qquad
    \subfloat[\centering Pre-impact state]{{\includegraphics[width=3cm]{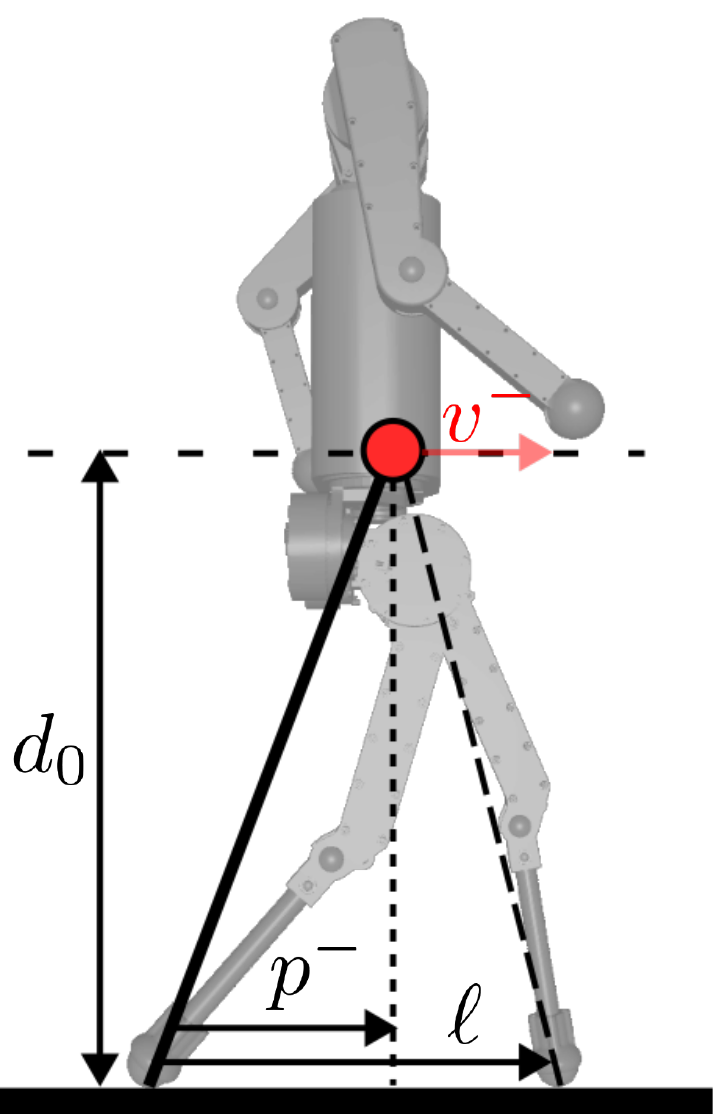} }}%
    \caption{The HLIP states $p$ and $v$ are defined as the linear horizontal COM movement relative to the stance foot.}%
    \label{fig:hlip_model}%
    \vspace{-1.4em}
\end{figure}

\subsection{Trajectory Optimization}\label{subsec:background:trajectory_optimization}
Given the dynamics and control objectives of the HZD framework, the next step is to generate reference trajectories for the outputs specified in the HZD setup \eqref{eq:output_constraint} via trajectory optimization. We use the FROST toolbox, which utilizes direct collocation, to generate the trajectories \cite{hereid2017frost}. The toolbox has previously been used to generate periodic gaits for various walking robots, herein quadrupeds \cite{ma2020coupled}, exoskeletons \cite{harib2018feedback},  and bipeds \cite{reher2019dynamic}.  

The problem of finding stable periodic gaits can be formulated as an optimization problem where it is easy to add additional constraints
\begin{align}
       \underset{\alpha, X}{\textrm{argmin}} & \ \Phi(X)
        \label{eq:standard_optimization_problem} \\
        \textrm{s.t.} \quad & \dot{x} = f_{cl}(x) \tag{Closed-Loop Dynamics} \\
        & \Delta(\mathcal{S} \cap \mathcal{Z}_\alpha) \subset \mathcal{Z}_\alpha \tag{Impact-Invariance}\\
        & X_{\min} \leq X \leq X_{\max}  \notag \tag{Decision Variables}\\
        & c_{\min} \leq c(X) \leq c_{\max} \tag{Physical Constraints}\\
        & a_{\min} \leq a(X) \leq a_{\max}.\tag{Essential Constraints} 
\end{align}
In \eqref{eq:standard_optimization_problem}, $X$ is the set of decision variables, $\Phi(X)$ is the cost function that determines what objective to minimize, e.g. energy usage per distance walked, maximum joint torque, etc.. The physical constraints are constraints that ensure that the obtained gait is physically feasible, such as friction cone constraints \cite{underactuated} and actuator constraint limits. The essential constraints \cite{tucker2021preference} are constraints that are used to shape the motion of the gait to achieve certain objectives. Examples include step length and step period constraints, walking speed constraints, and minimum swing foot height constraints.

\subsection{Hybrid Linear Inverted Pendulum Stabilization}\label{subsec:background:hlip}

Solving the optimization problem in \eqref{eq:standard_optimization_problem} is time consuming and must be done offline. To stabilize the nominal gaits online, we therefore utilize the HLIP model, which can easily be run in real-time. The HLIP model considers a planar linear inverted pendulum model where a point mass moves relative to a stance foot. The model has two states $p, \ v \in \mathbb{R}$ which are the horizontal COM position and velocity relative to the stance foot position, respectively. This can be seen in Fig. \ref{fig:hlip_model}. The model also assumes a fixed height $d_0 \in \mathbb{R}$. 


Let the two-dimensional pre-impact state used in the HLIP model be represented by
\begin{equation}
    \mathbf{z}
    =
    \begin{bmatrix}
        p^{-} \\
        v^{-}
    \end{bmatrix}
    \in
    \mathbb{R}^2.
\end{equation}
The goal of the HLIP controller is to control a robot through discrete pre-impact step-to-step (S2S) dynamics given by
\begin{equation}
    \label{eq:step2step_dynamics}
    \mathbf{z}_{k+1}^\textnormal{H} = \mathbf{A} \mathbf{z}_{k}^\textnormal{H} + \mathbf{B} \ell_{k}^\textnormal{H},
\end{equation}
where the pre-impact state at step $k+1$ is directly determined by the pre-impact state and step length at step $k$. The superscript $\textnormal{H}$ refers to the HLIP model. Analogously to \cite{xiong20223}, the matrices $\mathbf{A} \in \mathbb{R}^{2 \times 2}$ and $\mathbf{B} \in \mathbb{R}^{2 \times 1}$ are functions of the single support leg period, $T_\textnormal{SSP}$, and the double support leg period, $T_\textnormal{DSP}$. By choosing $T_\textnormal{SSP}$ and $T_\textnormal{DSP}$ to be constant, $\mathbf{A}$ and $\mathbf{B}$ become constant. Since the only control input to the S2S dynamics is the step length $\ell$, HLIP has to use the step length to stabilize the S2S dynamics.

Consider the reduced-order pre-impact state given by
\begin{equation}
    \mathbf{z}_{k+1}^\textnormal{R} = \mathcal{P}(\mathbf{q}_k, \dot{\mathbf{q}}_k, \tau(t)),
\end{equation}
where the superscript R refers to the robot. 

The robot's S2S dynamics are approximated by using HLIP's S2S dynamics and the discrepancy between the true robot dynamics and HLIP dynamics is treated as a disturbance $\mathbf{w}_k$
\begin{equation}
\label{eq:step2step_robot_dynamics}
\begin{split}
    \mathbf{z}_{k+1}^\textnormal{R} 
    = 
    \mathbf{A} \mathbf{z}_{k}^\textnormal{R} + \mathbf{B} \ell_{k}^\textnormal{R} + \mathbf{w}_k,
\end{split}
\end{equation}
where $\mathbf{w}_k$ is given by
\begin{equation}
    \mathbf{w}_k := \mathcal{P}(\mathbf{q}_k, \dot{\mathbf{q}}_k, \tau(t)) - \mathbf{A} \mathbf{z}_{k}^\textnormal{R} - \mathbf{B} \ell_{k}^\textnormal{R}.
\end{equation}

The goal then becomes to regulate the step length $\ell$ of the robot at the pre-impact state so that error S2S dynamics between the HLIP state and the robot's state become zero. That is, we want to regulate $\mathbf{e}:=\mathbf{z}^\textnormal{R} - \mathbf{z}^\textnormal{H} \in \mathbb{R}^2$ to zero. 

The discrete linear controller for the robot's S2S dynamics is given by
\begin{equation}
    \label{eq:hlip_step_controller}
    \ell_k^\textnormal{R} = \ell_k^\textnormal{H} + \mathbf{K}(\mathbf{z}_k^\textnormal{R} - \mathbf{z}_k^\textnormal{H}),
\end{equation}
where $\mathbf{K} \in \mathbb{R}^{1 \times 2}$ is a feedback gain matrix. Plugging this controller into \eqref{eq:step2step_robot_dynamics} results in the following closed-loop S2S error dynamics 
\begin{equation}
    \mathbf{e}_{k+1} = (\mathbf{A} + \mathbf{BK})\mathbf{e}_{k} + \mathbf{w}_k.
\end{equation}

To stabilize the discrete closed-loop S2S error dynamics, we have to select $\mathbf{K}$ such that all the eigenvalues $\lambda_i$ of the closed-loop system satisfies 
\begin{equation}
    \label{eq:eigen_value}
    |\lambda_{i}(\mathbf{A} + \mathbf{B}\mathbf{K})| < 1 , \ \textnormal{for} \ i \in \{1, 2\}.
\end{equation}

An example of a $\mathbf{K}$ achieving this is the deadbeat gain $\mathbf{K}_{\text{db}}$ which can be calculated from
\begin{equation}
    \label{eq:deadbeat}
    (\mathbf{A} + \mathbf{B}\mathbf{K}_{\text{db}})^2 = \mathbf{0}.
\end{equation}

%% file: 03_method.tex
\section{HZD Gait Stabilization Using an HLIP Regulator}\label{sec:gait_stabilization}

In this section we showcase our framework on how to control underactuated bipedal robots. 
The framework consists of two parts. First, the nominal output trajectories are generated using full-order model based HZD trajectory optimization, incorporating the reduced-order HLIP model as a constraint within the gait generation. This ensures alignment between the full-order gait and the reduced-order model. Next, we demonstrate how to modify the nominal HZD trajectories by using HLIP as a footstep regulator, in order to stabilize the underactuated COM states online via S2S dynamics. 

\subsection{Gait Generation with HLIP Constraints}

As explained in section \ref{subsec:background:hlip}, HLIP stabilization is concerned with controlling the planar underactuated horizontal COM state to a specific pre-impact state by regulating the step length. Because the HLIP model uses the S2S dynamics, it is important that the pre-impact state of the HLIP model coincides with the pre-impact state of the HZD gait. This ensures that the HLIP regulator will attempt to stabilize the robot to the same periodic orbit as the HZD gait.

Calculation of the HLIP pre-impact state depends on the chosen orbit characterization, which is a notion used in the HLIP model to quantify how many steps that are needed to stabilize the HLIP model back to its nominal state \cite{xiong20223}. The two most common choices are period-1 orbits and period-2 orbits, which stabilizes the HLIP model back its nominal pre-impact state after 1 and 2 steps respectively. For sagittal walking, which is left and right symmetric, we choose to use period-1 walking. It can be shown that for a desired walking speed and a fixed step period, the period-1 orbit is fully characterized and the pre-impact state, $\mathbf{z}$, can be calculated directly with a unique solution \cite{xiong20223}. For the left to right frontal plane control, a period-2 orbit is used. To fully characterize the period-2 orbit, a step width must also be specified, in addition to the lateral velocity and step period \cite{xiong20223}. This allows us to choose an offset between the nominal stance foot position for the left and right foot, so that we can avoid collisions between the legs. Therefore, for a given desired sagittal and frontal velocity, a fixed step period, and a fixed desired step width, the pre-impact states in both the sagittal plane and the frontal plane can be computed with closed form solutions.

We then have that for a desired gait velocity, the pre-impact states and therefore HLIP states throughout the gait can be computed. We can then ensure that the horizontal COM position and velocity follow the HLIP states throughout the gait by enforcing the following constraints 
\begin{equation}
    \begin{bmatrix}
        p^{\mathrm{R}}_{\mathrm{COM}, x/y}(X) \\
        v^{\mathrm{R}}_{\mathrm{COM}, x/y}(X) \\
    \end{bmatrix}
    =
    \begin{bmatrix}
        p_{x/y}^{\mathrm{H}}(X) \\
        v_{x/y}^{\mathrm{H}}(X) \\
    \end{bmatrix}.
    \label{eq:hlip_constraints}
\end{equation}
Adding the HLIP constraints in \eqref{eq:hlip_constraints} to the standard HZD gait optimization problem results in the modified HLIP consistent HZD gait optimization problem given by
\begin{align}
       \underset{\alpha, X}{\textrm{argmin}} & \ \Phi(X)
        \label{eq:modified_optimization_problem} \\
        \textrm{s.t.} 
        \quad & \textnormal{Constraints from \eqref{eq:standard_optimization_problem}} \notag \\
        & \textnormal{Constraints from \eqref{eq:hlip_constraints}}. \notag
\end{align}
When solving the trajectory optimization problem we choose the following set of outputs:
\begin{equation}
    \mathcal{Y} 
    = 
    [
    p_{\textnormal{COM},z} 
    \quad \Phi_\textnormal{torso}^\top
    \quad \mathbf{p}_\textnormal{swf}^\top
    \quad \mathcal{Y}_\mathrm{other}^\top 
    ]^\top,
    \label{eq:trajopt_outputs}
\end{equation}
where $p_{\textnormal{COM},z}$ is the COM height of the robot, $\Phi_\textnormal{torso}$ is the 3D orientation of the robot's torso, $\mathbf{p}_\textnormal{swf}$ is the 3D swing foot position of the robot, and $\mathcal{Y}_\mathrm{other}$ are other outputs of the robotic system, for example arm trajectories. The minimum required outputs necessary to regulate the robot with our approach are the COM height, COM roll and pitch, and the 3D swing foot position. The COM yaw and other additional outputs may be added depending on the actuated degrees of freedom of the robot. All the outputs are represented using Bézier polynomials. In addition to the outputs, the trajectory optimization on the full-order system  also generates trajectories for the underactuated COM position states which we here denote as $\mathcal{Z} =[x_\textnormal{COM} \quad y_\textnormal{COM}]^\top.$

\begin{figure}[!t]
    \centering
    \includegraphics[width=\linewidth]{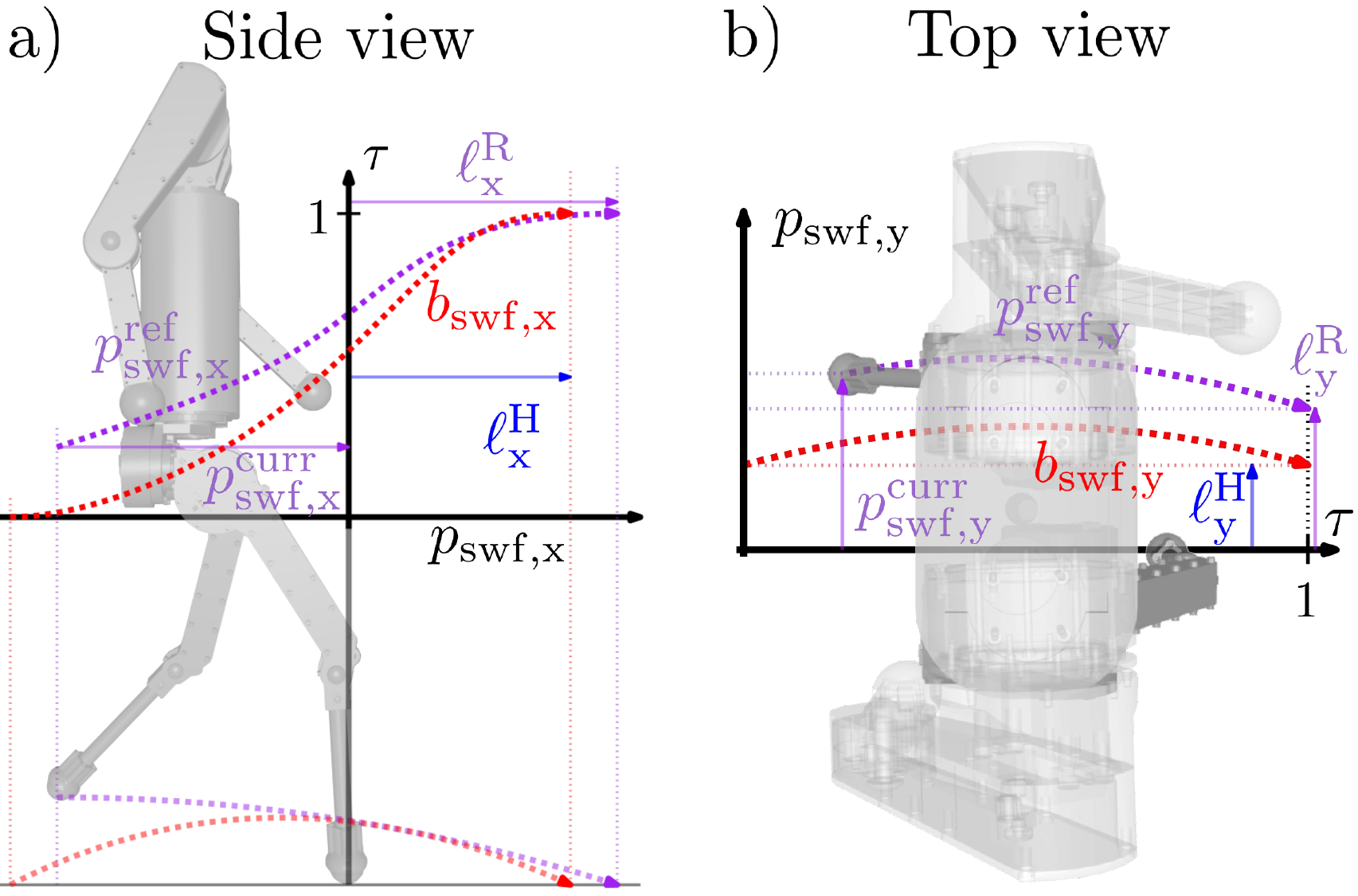}
    \caption{The swing foot $x$ and $y$ position trajectories are functions of the current swing foot position, the desired step lengths, and the nominal HZD gait swing foot trajectories.}
    \label{fig:blending}
    \vspace{-1.8em}
\end{figure}

\subsection{Modifying the Nominal HZD Outputs}
When performing periodic locomotion, it is in theory, sufficient to track the output trajectories generated in \eqref{eq:trajopt_outputs}. However, verifying the stability of the zero dynamics is hard for 3D walking and
some level of model discrepancy and tracking error is bound to exist. The underactuated states from $\mathcal{Z}$ will therefore, in general, deviate from their nominal periodic states and eventually cause the robot to fall over. To account for this, we use an HLIP controller as a step length regulator to stabilize the underactuated states, i.e. the COM $x$ and $y$ position. We use robot's actuators to control the outputs of the robot, and stabilize the remaining underactuated dynamics through S2S control. 

As described in section \ref{subsec:background:hlip}, the HLIP regulator can stabilize the planar horizontal COM state of the robot by regulating the step length of the robot according to \eqref{eq:hlip_step_controller}. Since the HLIP model is defined in the plane, we employ two sets of HLIP regulators, one in the sagittal plane and one in the frontal plane, which are used to control the swing foot $x$ and $y$ position, respectively. 

The nominal horizontal swing foot trajectories from the optimization problem are generated as Bézier polynomials given by $b_\mathrm{swf,x}(\tau)$ and $b_\mathrm{swf,y}(\tau)$. To accommodate for tracking errors and varying step lengths based on the HLIP regulators, it is necessary to generalize the swing foot $x$/$y$ trajectories. The period-1 horizontal swing foot trajectory can be normalized through the following transformation:
\begin{equation}
    \bar{b}_\mathrm{swf,x}(\tau)
    =
    \frac{b_\mathrm{swf,x}(\tau) - b_\mathrm{swf,x}(0)}
    {b_\mathrm{swf,x}(1) - b_\mathrm{swf,x}(0)},
    \label{eq:swf_x_modified}
\end{equation}

and the period 2 horizontal swing foot trajectory can be normalized through the following transformation:
\begin{equation}
    \bar{b}_\mathrm{swf,y}(\tau)
    =
    \frac{b_\mathrm{swf,y}(\tau)}
    {b_\mathrm{swf,y}(0)}.
    \label{eq:swf_y_modified}
\end{equation}

The normalization in \eqref{eq:swf_x_modified} can then be used to calculate the sagittal swing foot x trajectory through:
\begin{align}
    \label{eq:swf_x_ref_theory}
   {p_\textnormal{swf,x}^\textnormal{ref}}(\tau) 
    &= (1 - \tau) p_\textnormal{swf,x}^{\textnormal{curr}} \nonumber \\
    &+ \tau ([1 - \bar{b}_\textnormal{swf,x}(\tau)] b_\textnormal{swf,x}(0) + \bar{b}_\textnormal{swf,x}(\tau) \ell_\textnormal{x}^\textnormal{R}),
\end{align}

and \eqref{eq:swf_y_modified} can be used to calculate the frontal swing foot y trajectory through:
\begin{equation}
    \label{eq:swf_y_ref_theory}
    {p_\textnormal{swf,y}^\textnormal{ref}}(\tau) 
    = 
    (1 - \tau) p_\textnormal{swf,y}^\textnormal{curr} + \tau \bar{b}_\mathrm{swf,y}(\tau) \ell_\textnormal{y}^\textnormal{R}. \\
\end{equation}

Figure \ref{fig:blending} illustrates the online swing foot trajectory generation from \eqref{eq:swf_x_ref_theory} and \eqref{eq:swf_y_ref_theory}, based on the current swing foot position, the desired step lengths, and the nominal HZD trajectories. 
The task space trajectories from the HZD gait generation in \eqref{eq:trajopt_outputs} together with the modified swing foot trajectories can be tracked through a task space controller.


%% file: 04_application.tex
\begin{figure}[!t]
    \centering
    \includegraphics[width=8.6cm]{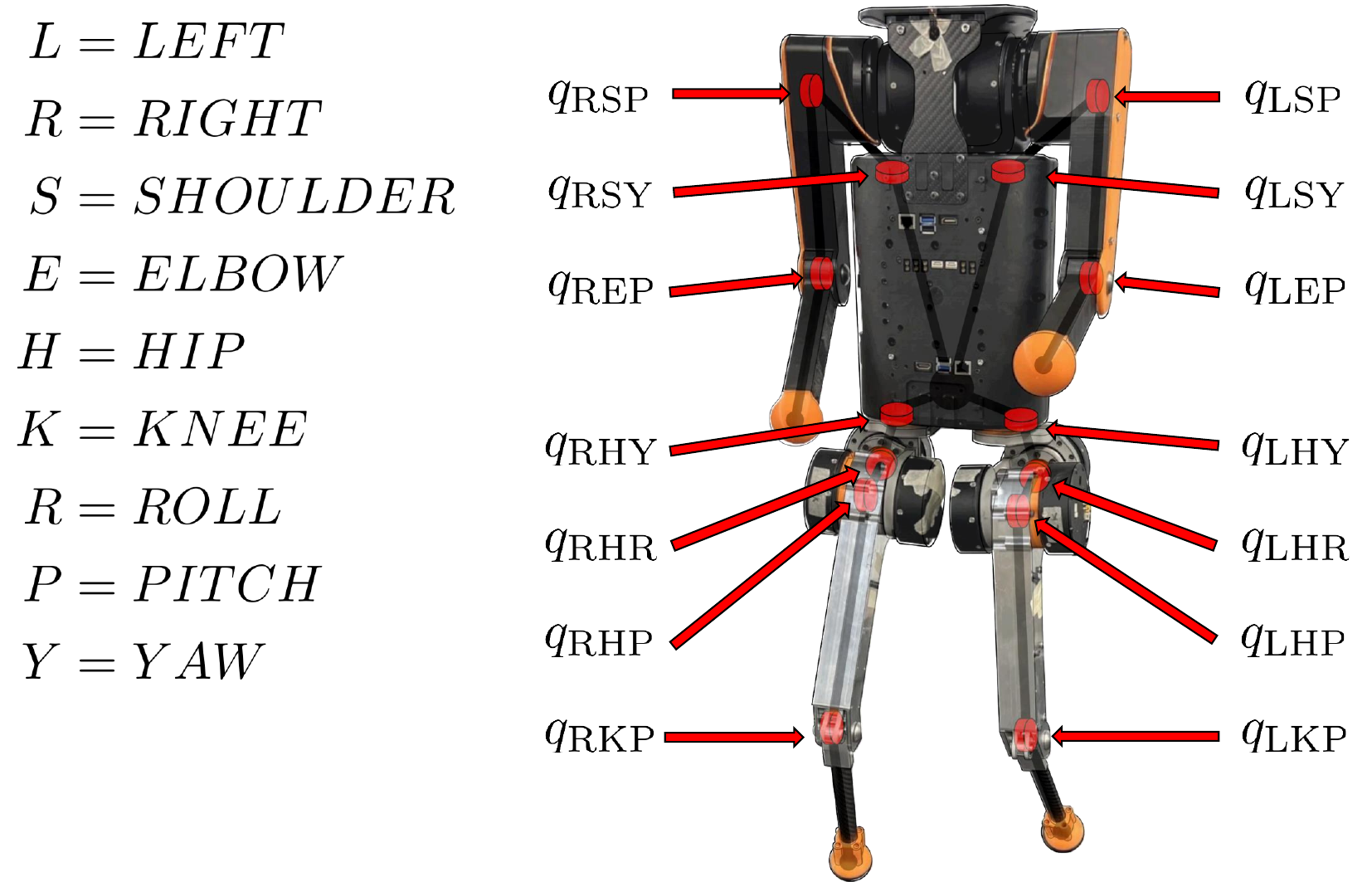}
    \caption{The joint coordinates of the humanoid robot ADAM.}
    \label{fig:adam_platform}
    \vspace{-1.8em}
\end{figure}
\section{Application to a Humanoid Robot}\label{sec:application}

To evaluate our control framework, we conduct a series of tests in both simulation and on hardware with the humanoid robot ADAM which was developed in \cite{ghansah2023humanoid}. ADAM is a 20 DoF point foot humanoid robot, standing $0.95$ m tall with a total weight of $15$ kg. Each leg contains 4 actuators, which are connected in series, where the innermost controls the hip yaw joint, the second controls the hip roll joint, and the last two control the hip pitch and knee pitch joint, respectively. Each arm contains three actuators, where the two innermost control the yaw and pitch joint of the shoulder, and the last actuator controls the elbow pitch joint. The full joint configuration of the robot can be seen in Fig. \ref{fig:adam_platform}.

\subsection{Generating Reference Trajectories}

The HLIP consistent HZD reference trajectories were computed offline by using FROST. Gaits were created for multiple walking speeds and then collected into a gait library. For each desired walking speed we fixed both the desired stepping period and step length to simplify the gait generation and to ensure consistent gait periods for all the gaits. Since the walking speed was fixed for each gait, the HLIP constraints from \eqref{eq:hlip_constraints} could easily be added as lower and upper bounds to the COM $x$ and $y$. Furthermore, to satisfy the fixed height constraint of the HLIP model, we also add tight lower and upper bounds on the COM $z$ height. In addition to the minimal set of outputs, we also generate outputs for the torso yaw and arm joints. When generating new gaits to the gait library, the last generated gait is used as the initial guess for the next one in order to achieve more consistency between the gaits. For the cost function, a combination of cost of transport and torque squared minimization is used.

\begin{figure*}[t!]
    \centering
  \includegraphics[width=17.5cm, height=8.27cm]{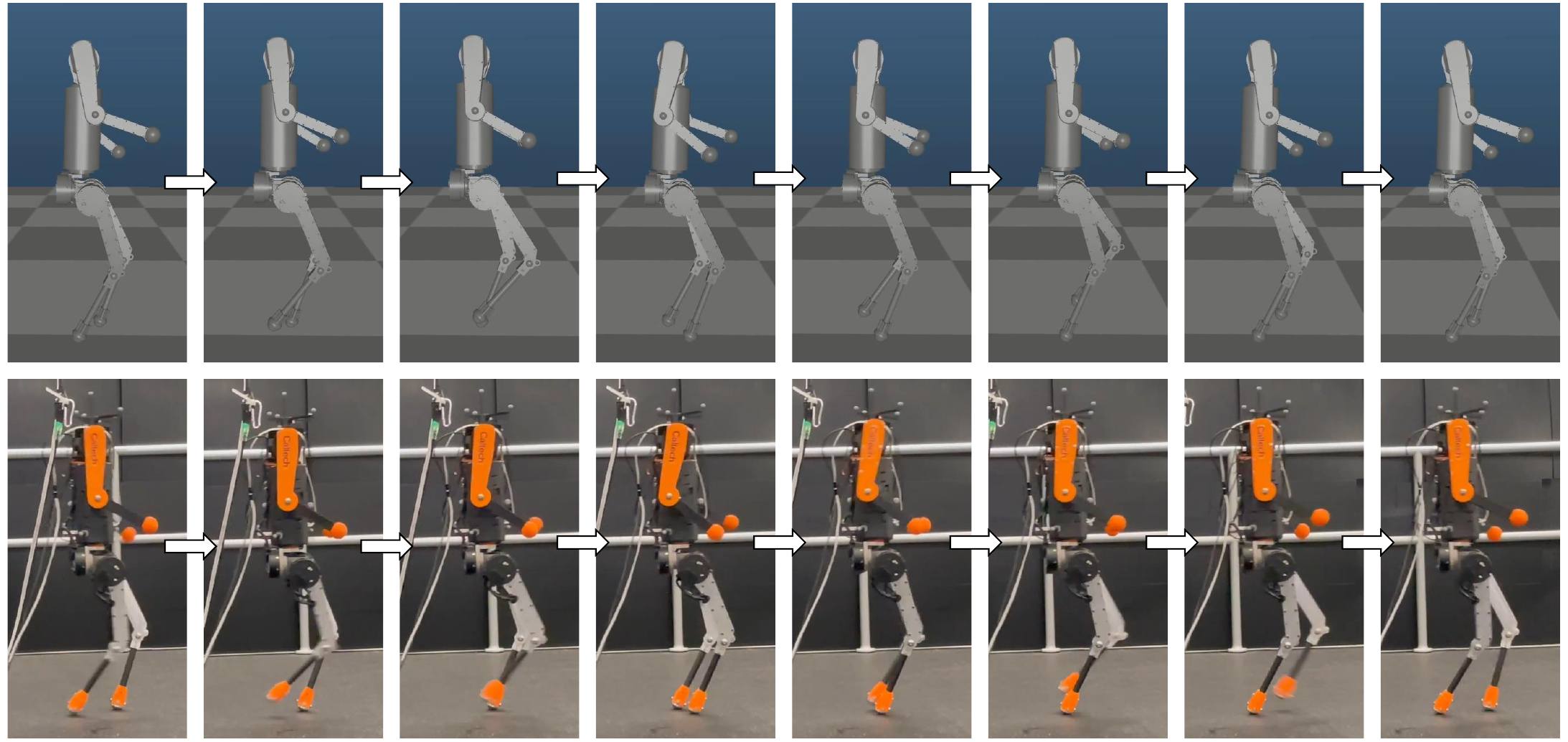}
  \caption{The upper sequence shows a nominal HZD gait in simulation, while the lower one shows the hardware realization.}
  \label{fig:gait_tiles}
  \vspace{-1.0em}
\end{figure*}

\begin{figure}[t!]
    \centering
  \includegraphics[width=0.9\linewidth]{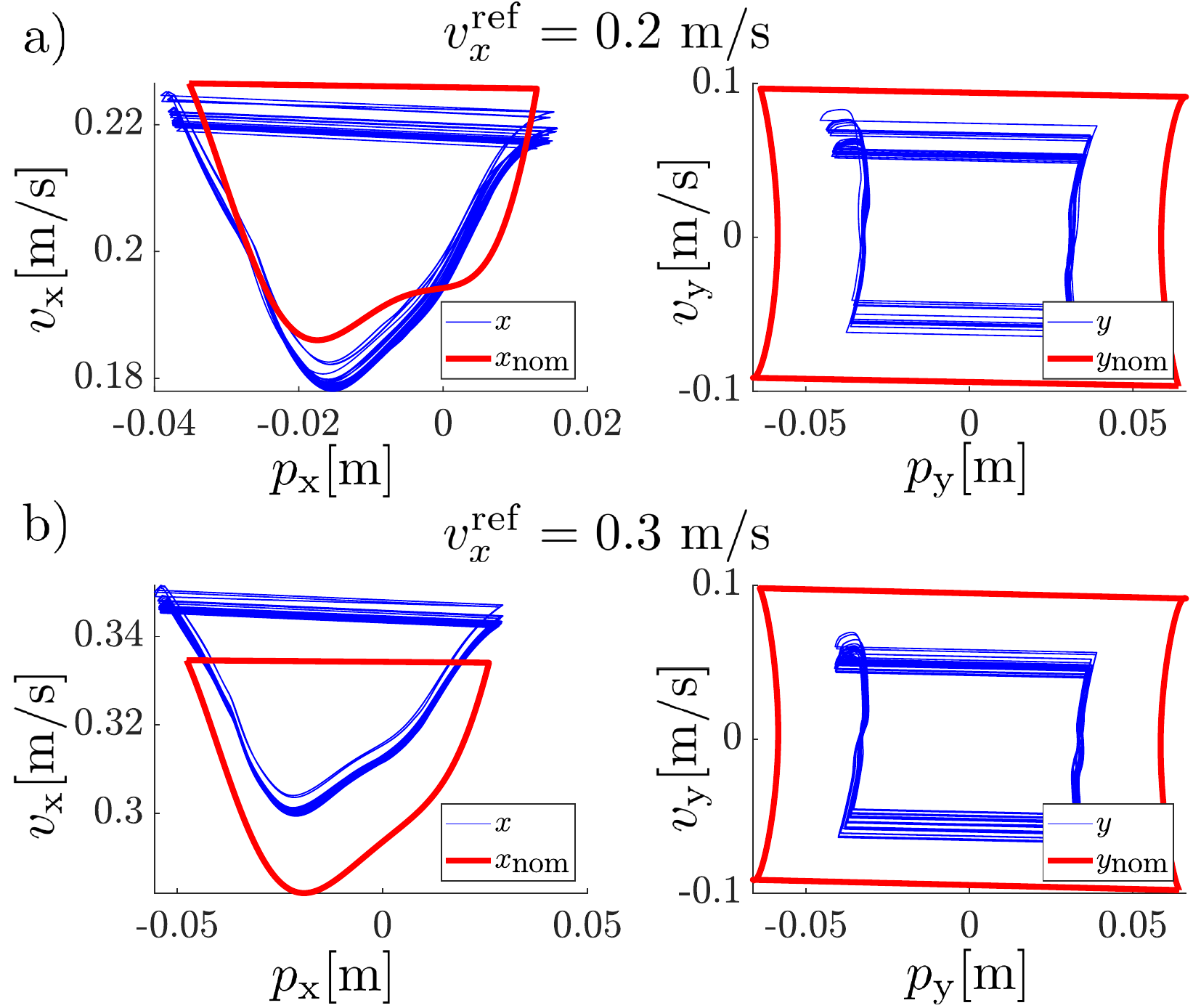}
  \caption{Phase portraits of the simulated robot's underactuated COM x and y states over several steps are shown together with the orbits from the nominal HZD gaits. a) phase portraits for reference velocity $v_x^{\text{ref}} = 0.2$ m/s. b) phase portraits for reference velocity $v_x^{\text{ref}} = 0.3$ m/s.}
  \label{fig:phase_portrait}
  \vspace{-1.5em}
\end{figure}

\begin{figure}[t]
    \centering
  \includegraphics[width=\linewidth]{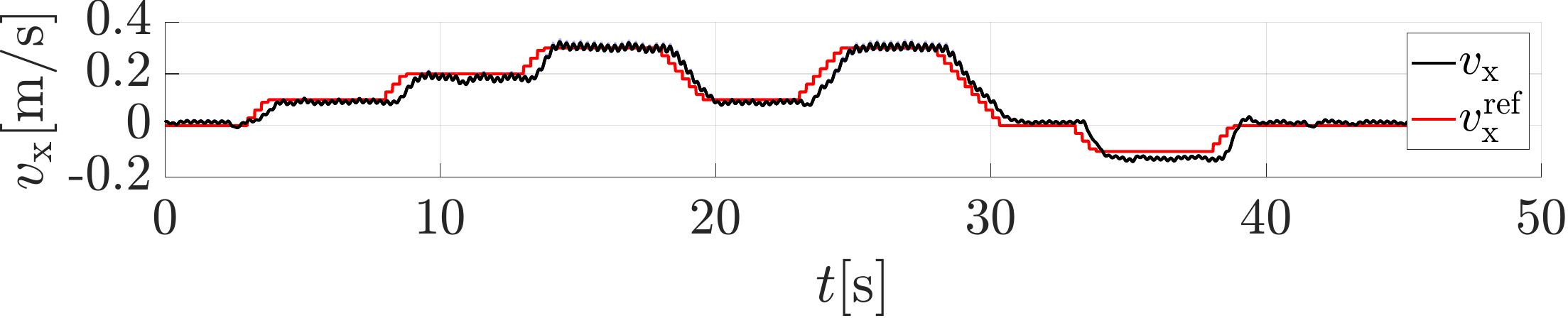}
  \includegraphics[width=\linewidth]{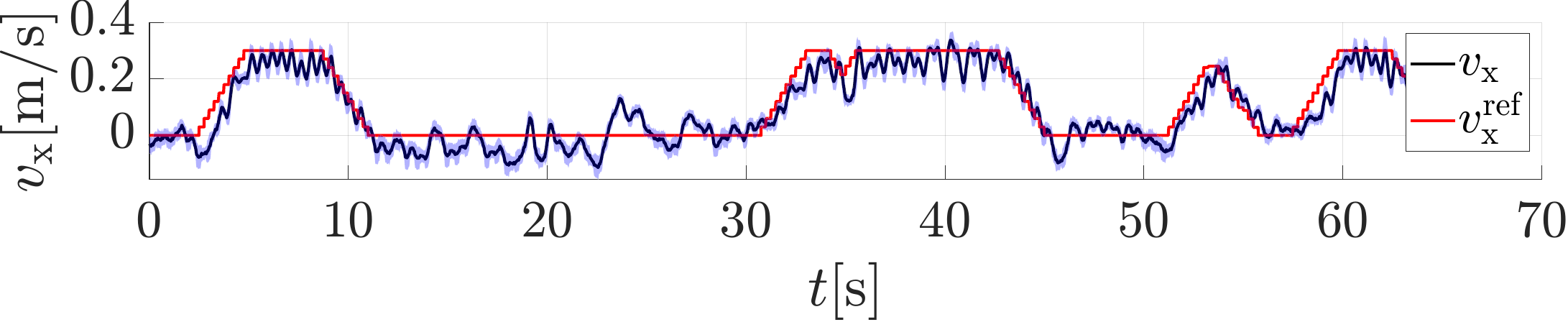}  \caption{Tracking a desired COM x velocity in simulation (top) and on hardware (bottom).   
  The filtered state is calculated by using the moving average over one step period, 0.25 seconds. The blue bounds quantify the 1-standard deviation variation over one step period.  On hardware, the reference signal was obtained via joystick control.}
  \label{fig:simulation_velocity_tracking}
  \label{fig:hardware_velocity_tracking}
  \vspace{-1.5em}
\end{figure}

\subsection{Controller Setup}

When commanding the robot to walk at a certain speed, a convex combination of two sets of outputs corresponding to the gaits with the closest reference speeds from the gait library are used. All of the nominal outputs from the gait generation can be used as direct reference trajectories apart from the swing foot $x$ and $y$ trajectories, where we use \eqref{eq:swf_x_ref_theory} and \eqref{eq:swf_y_ref_theory} instead. To determine the step length in the x and y direction we refer to \eqref{eq:hlip_step_controller}. For a fixed walking speed, $\mathbf{z}_k^\mathrm{H}$ and $\ell_k^\mathrm{H}$ are given, and the pre-impact state of the robot, $\mathbf{z}_k^\mathrm{R}$, has to be estimated. We estimate $\mathbf{z}_k^\mathrm{R}$ from the robot's current COM state relative to its stance foot, the predicted time to impact, and by assuming that the robot moves like an inverted pendulum. To avoid excessive oscillations in the horizontal swing foot trajectories, we also feed the calculated step lengths through a low-pass filter. The task space control outputs are mapped into low-level joint commands through a inverse kinematics PD controller.


%% file: 05_experiments.tex
\begin{figure*}[t!]
    \centering
  \includegraphics[width=\linewidth]{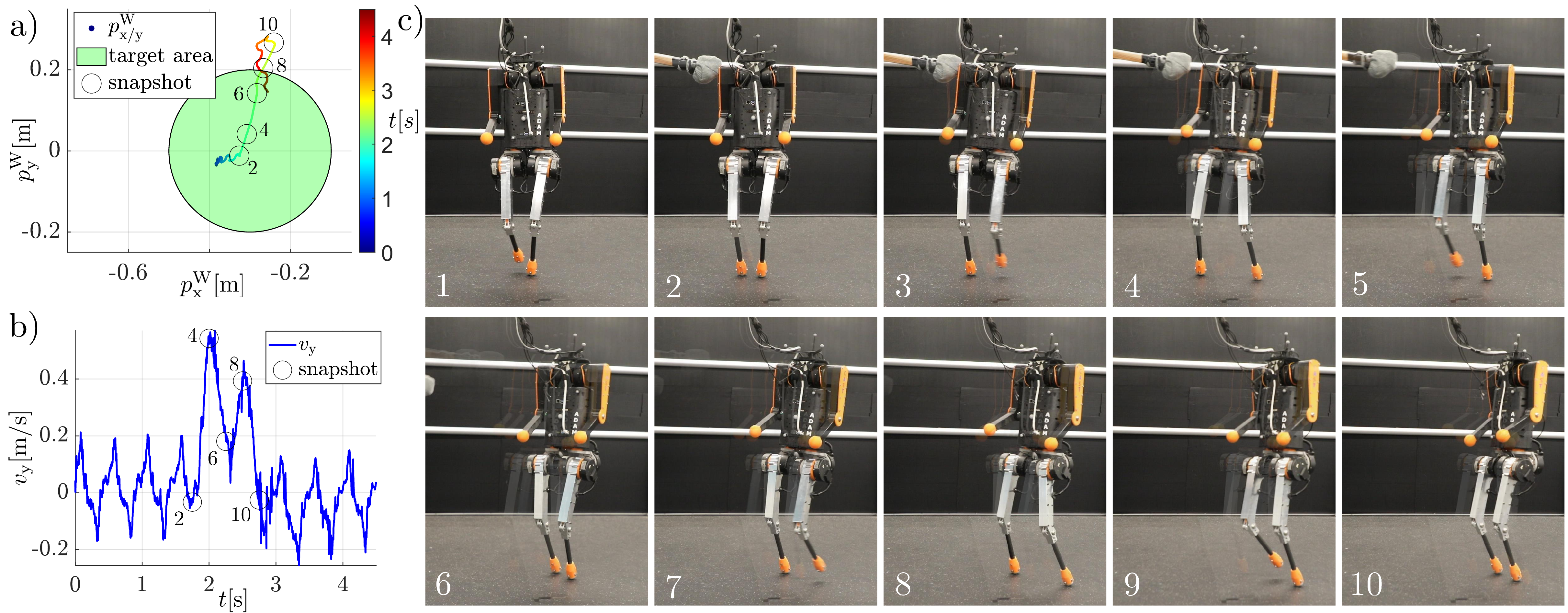}
  \caption{The robot is able to stabilize itself after being perturbed by an external disturbance. The external push causes a large increase in the lateral velocity and the robot is able to stabilize itself in 4 steps before it reaches its nominal period orbit.}
  \label{fig:push_recovery_experiment}
  \vspace{-1.5em}
\end{figure*}

\section{Experimental Validation}\label{sec:experimental}

This section presents the validation of the proposed approach, both in simulation and experimentally.

\subsection{Simulation Experiments}\label{sec:exeriments:simulation}

To evaluate the performance of our control framework in simulation, the physics based engine MuJoCo \cite{todorov2012mujoco} is used (see Fig. \ref{fig:gait_tiles}). In both simulation and on hardware we use a constant step period of 0.25 s. The robot is capable of walking at a maximum forward speed of $0.5$ m/s, and can also walk backwards, sideways, and turn. Figures \ref{fig:phase_portrait}a and \ref{fig:phase_portrait}b show phase portraits for the underactuated COM x and y position when walking at 0.2 and 0.3 m/s, respectively. In both cases, it is clear that the controller is able to stabilize the robot, which is reflected by the periodic behavior of the phase portraits. Figure \ref{fig:phase_portrait} also shows that the underactuated states converge to stable periodic orbits that resemble the nominal periodic orbits from the generated HZD gaits.  
Additionally, Fig. \ref{fig:simulation_velocity_tracking} shows that the controller enables the system to track COM velocity commands in the sagittal plane. The robot is able to track the desired velocity accurately at different velocities and also handles the transitions between them.



\begin{figure*}[t]
    \centering
  \includegraphics[width=17.5cm, height=5.7cm]{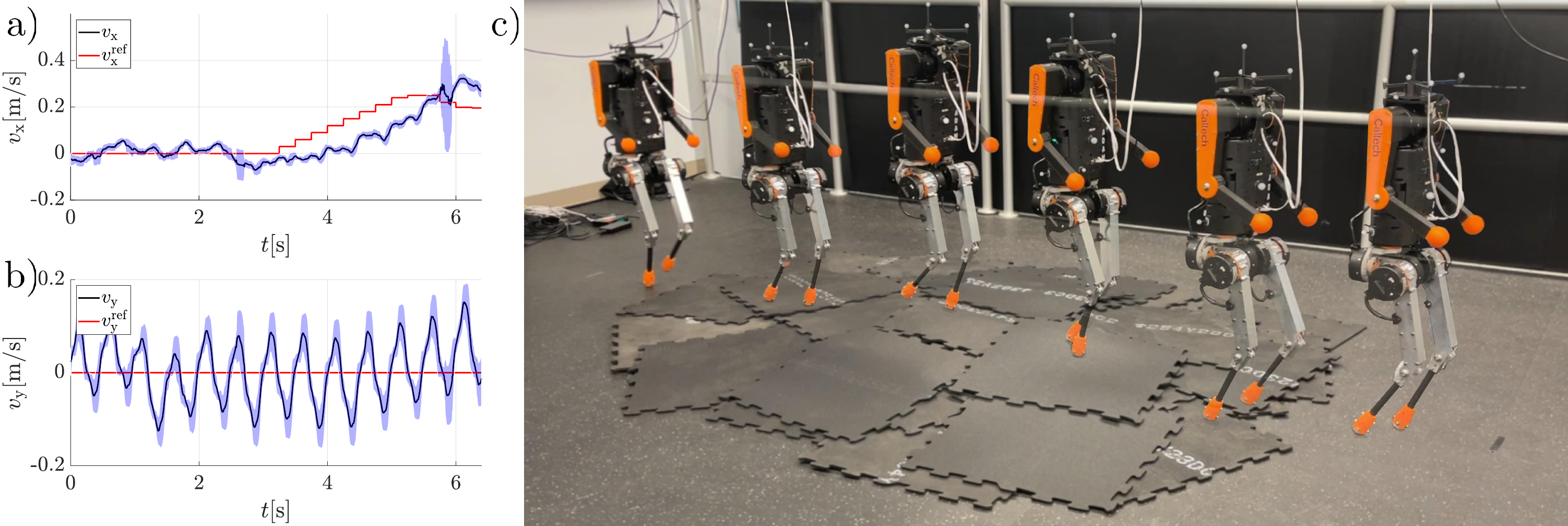}
  \caption{The robot is able to traverse unknown unstructured terrain. Here it walked over a pile of rubber mats with a maximum height of 0.11 m. The reference velocity is shown in red, the filtered average velocity over half a step period is shown in black, with the shaded-area indicating the one standard deviation variation over half a step period. a) Velocity $x$ tracking. b) Velocity $y$ tracking. c) Snapshots of the rough terrain walking.}
  \label{fig:uneven_terrain}
  \vspace{-1.5em}
\end{figure*}

\subsection{Hardware Experiments}\label{sec:experiments:hardware}
For the hardware experiments, the state estimation and control of the robot was done on an off-board computer which was connected to the robot over Ethernet. The computer was equipped with a 12th Gen Intel i7-12800H 2.40 GHz processor and 16 GB of RAM. More specifically for the state estimation, a VectorNav VN-100 IMU and Optitrack Motion Capture System were used to provide orientation, angular velocity, linear acceleration, and velocity measurements. The measurements were fed into an indirect Kalman filter \cite{fossen1999guidance} to estimate the states.

The experiments are composed of three different setups: nominal walking experiments, push recovery experiments, where the robot has to stabilize itself after being perturbed by external forces, and rough terrain walking experiments, where the robot has to traverse unstructured, uneven terrain.

\subsubsection{Nominal Walking} Similarly, as in simulation, the real robot was able to walk at different speeds and track various velocity commands. Figure \ref{fig:hardware_velocity_tracking} shows the robot's velocity tracking abilities when controlling it around using a joystick. Despite the oscillating behavior of the robot, we see that the robot is able to follow the commanded velocity commands. Realization of a nominal HZD gait from the optimization problem on hardware is displayed in Fig. \ref{fig:gait_tiles}.


\subsubsection{Push Recovery}

To test the robot's robustness against external disturbances, we applied external forces to the robot while it was walking. In Fig. \ref{fig:push_recovery_experiment}, we ran an experiment where the robot was tasked with staying inside a target area. While walking in place, the robot is exposed to a push from the side, which causes a large velocity in the lateral direction. The large velocity induced by the external disturbance is shown in Fig. \ref{fig:push_recovery_experiment}b), where it suddenly increases, forcing the robot to move to the side. Figure \ref{fig:push_recovery_experiment}c) illustrates how the robot adjusts the step width to reduce the lateral velocity. Figure \ref{fig:push_recovery_experiment}a) shows how the robot starts within the target area, then gets pushed outside, and after stabilizing itself, it walks back into the target area.

\subsubsection{Rough Terrain}

To test the controller's stability when walking on unknown, unstructured terrain, we performed several experiments where the robot had to traverse a pile of rubber mats with a maximum height of $0.11$ m. 
Figure \ref{fig:uneven_terrain}a) and \ref{fig:uneven_terrain}b) show the x and y directional velocity tracking capabilities, respectively, while walking on the rough terrain. 
Even though the velocity x tracking is degraded compared to the nominal walking in Fig. \ref{fig:simulation_velocity_tracking}, the robot is still able to follow the commanded velocity.
The successful traversing over the obstacles is visualized in Fig. \ref{fig:uneven_terrain}c). 

\subsection{Discussion}
Despite the underactuation caused by the point feet, the proposed framework still enables robust locomotion of the robot in both simulation and on hardware. However, we faced challenges associated with yaw slippage of the stance foot, as we used rigid point feet made out of metal which provided minimal yaw friction. During the experiments, maintaining the constant robot height and fixed orientation contributed to mitigating the yaw slippage. Introducing the feet with multiple contact points is anticipated to further prevent this slippage. A video of the experiments is available in \cite{video}.

%% file: 06_conclusion.tex
\section{Conclusion}\label{sec:conclusion}
In this paper we present a control framework to stabilize highly underactuated bipedal robots by generating nominal reference trajectories using the HZD framework, while incorporating HLIP dynamics into the gait generation. 
Based on the nominal reference trajectories, an HLIP regulator stabilizes the HZD gaits online by modifying the nominal swing foot trajectories. The control framework was tested extensively in both simulation and on hardware on the point foot humanoid robot ADAM. We demonstrated the robustness of the proposed framework by achieving robust locomotion subject to various unknown external disturbances and by traversing uneven terrain.
%
In the near future, we intend to introduce the feet to prevent slippage at the contact point and to improve the locomotion performance of the highly underactuated humanoid system. Moreover, we plan to explore the advantages of utilizing actuated ankles. 